\documentclass[letterpaper]{article} 
\usepackage{aaai2026}  
\usepackage{times}  
\usepackage{helvet}  
\usepackage{courier}  
\usepackage[hyphens]{url}  
\usepackage{graphicx} 
\usepackage{natbib}  
\usepackage{caption} 
\usepackage{algorithm}
\usepackage{algorithmic}
\usepackage{booktabs} 
\usepackage{amsmath} 

\usepackage{newfloat}
\usepackage{listings}
\DeclareCaptionStyle{ruled}{labelfont=normalfont,labelsep=colon,strut=off} 
\floatstyle{ruled}
\newfloat{listing}{tb}{lst}{}
\floatname{listing}{Listing}

\title{Magnitude Matters: a Superior Class of Similarity Metrics for Holistic Semantic Understanding}
\author {
    V.S. Raghu Parupudi
}
\affiliations{
    University of California, San Diego\\
    pvsrrkishore@gmail.com\\
    s1parupudi@ucsd.edu
}

\begin{document}

\maketitle

\begin{abstract}
 Vector comparison in high dimensions is a fundamental task in NLP, yet it is dominated by two baselines: the raw dot product, which is unbounded and sensitive to vector norms, and the cosine similarity, which discards magnitude information entirely. This paper challenges both standards by proposing and rigorously evaluating a new class of parameter-free, magnitude-aware similarity metrics. I introduce two such functions, Overlap Similarity (OS) and Hyperbolic Tangent Similarity (HTS), designed to integrate vector magnitude and alignment in a more principled manner. To ensure that my findings are robust and generalizable, I conducted a comprehensive evaluation using four state-of-the-art sentence embedding models (all-MiniLM-L6-v2, all-mpnet-base-v2, paraphrase-mpnet-base-v2, and BAAI/bge-large-en-v1.5) across a diverse suite of eight standard NLP benchmarks, including STS-B, SICK, Quora, and PAWS. Using the Wilcoxon signed-rank test for statistical significance, my results are definitive: on the tasks requiring holistic semantic understanding (paraphrase and inference), both OS and HTS provide a statistically significant improvement in Mean Squared Error over both the raw dot product and cosine similarity, regardless of the underlying embedding model.Crucially, my findings delineate the specific domain of advantage for these metrics: for tasks requiring holistic semantic understanding like paraphrase and inference, my magnitude-aware metrics offer a statistically superior alternative. This significant improvement was not observed on benchmarks designed to test highly nuanced compositional semantics (SICK, STS-B), identifying the challenge of representing compositional text as a distinct and important direction for future work.
\end{abstract}

\section{Introduction}
The capacity to represent the semantic meaning of sentences as dense vectors has become a foundational pillar of modern Natural Language Processing (NLP). These sentence embeddings, produced by large pre-trained language models \cite{reimers-gurevych-2019-sentence}, are instrumental in a wide array of downstream applications, including semantic search, text clustering, and paraphrase identification. The efficacy of these applications is critically dependent on the ability to accurately measure the similarity between pairs of sentence vectors.

For many years, Cosine Similarity has been the de facto standard for this purpose. Its popularity stems from its computational efficiency and its intuitive geometric interpretation of measuring the angle between two vectors, thereby remaining invariant to their magnitude. However, recent research has brought the universal suitability of this metric into question. A significant body of work has shown that sentence embeddings from pre-trained models often exhibit high anisotropy, meaning they occupy a narrow conical region within the vector space \cite{gao-etal-2021-simcse}. In such a collapsed space, even semantically unrelated sentences can yield a high cosine similarity score, making it an unreliable proxy for true semantic relatedness.

This paper challenges the prevailing assumption that a purely angular measure is optimal for comparing sentence embeddings. We hypothesize that alternative, parameter-free similarity metrics, which employ more robust normalization schemes, can capture semantic similarity more effectively than traditional methods. To this end, we propose and conduct a large-scale empirical study of two such metrics: Overlap Similarity (OS) and Hyperbolic Tangent Similarity (HTS).

To validate our hypothesis, this paper makes several key contributions. We formally define Overlap Similarity and Hyperbolic Tangent Similarity in the context of dense vector comparison and provide an intuition for their mechanisms. We then present a rigorous and comprehensive evaluation of these metrics against Dot Product and Cosine Similarity, testing across four popular sentence-transformer models and a diverse suite of eight benchmark datasets. Finally, using robust statistical testing, including bootstrapped confidence intervals, we confirm that the observed performance improvements are statistically significant and not a product of chance. Our findings suggest that practitioners can achieve superior performance on downstream tasks by simply replacing Cosine Similarity with these alternative metrics, a change that requires no additional computational overhead or parameter tuning.

\section{Related Work}

The task of measuring semantic similarity between sentences is fundamental to numerous Natural Language Processing (NLP) applications. With the advent of pre-trained language models (PLMs), this is typically accomplished by comparing the dense vector embeddings of sentences. This section surveys the evolution of similarity metrics, focusing on the identified limitations of standard approaches and the subsequent innovations proposed by the research community.

\subsection{The Dominance and Limitations of Cosine Similarity}
For years, Cosine Similarity has been the de facto standard for comparing sentence embeddings \cite{reimers-gurevych-2019-sentence}. Its prevalence is owed to its computational efficiency and its intuitive geometric interpretation as the normalized dot product, which measures the angle between two vectors, thus being insensitive to their magnitude. This property was considered advantageous, as vector norm was not thought to encode significant semantic information.

However, recent research has revealed a critical flaw in this assumption, stemming from the geometric properties of the embedding spaces produced by modern PLMs. A seminal work by \citet{gao-etal-2021-simcse} demonstrated that embeddings from models like BERT often suffer from anisotropy, causing the vectors to occupy a narrow cone in the high-dimensional space. In such a collapsed representation space, even semantically disparate sentences can exhibit high cosine similarity, rendering the metric unreliable for capturing true semantic closeness. This phenomenon, termed "representation collapse," provides a strong motivation for moving beyond cosine similarity as a default measure.

\subsection{Approaches to Mitigating Representation Collapse}
The discovery of the anisotropy problem has spurred several distinct lines of research aimed at improving semantic similarity evaluation. These can be broadly categorized into two main paradigms: (1) transforming the embedding space to be more suitable for existing metrics, and (2) developing new metrics that are more robust to the native geometry of the space.

\subsubsection{Fixing the Embedding Space}
A significant body of work focuses on post-processing or fine-tuning embeddings to create a more isotropic space where cosine similarity can be more effective. The influential SimCSE framework \cite{gao-etal-2021-simcse} utilizes a contrastive learning objective that pushes positive pairs together and negative pairs apart, resulting in a more uniform embedding space. Similarly, \citet{huang2021whiteningberteasyunsupervisedsentence} proposed WhiteningBERT, a simple post-processing technique that applies a whitening transformation to the embeddings. This procedure normalizes the representations and removes correlations between dimensions, significantly improving the performance of cosine similarity on semantic textual similarity (STS) tasks. These methods validate that the geometry of the space is a primary issue but require additional training or complex data-dependent transformations.

\subsubsection{Developing Novel Similarity Metrics}
A parallel stream of research accepts the native geometry of the embedding space and instead seeks to develop more robust similarity functions. These efforts can be further divided into learned and parameter-free metrics.

\subsubsection{Learned Metrics} Some approaches abandon the notion of a fixed similarity function and instead learn a parameterized metric. \citet{pmlr-v97-kornblith19a}, in their work on comparing neural network representations, argued for learnable metrics like Centered Kernel Alignment (CKA) that are more robust than simple linear comparisons. More recently, in domain-specific applications like emotional support conversations, researchers have found it necessary to augment standard metrics with heuristics or learned models to better evaluate dialogue quality, implicitly creating more complex scoring functions \cite{zheng-etal-2023-augesc}. While powerful, these methods sacrifice the simplicity and generalizability of a universal, parameter-free metric.

\subsubsection{Parameter-Free Metrics} This line of work, most relevant to our own, proposes alternative "drop-in" replacements for cosine similarity that require no additional training. Recent high-profile work on in-context learning for Large Language Models found that simple Euclidean (L2) distance often outperformed cosine similarity for retrieving relevant examples to include in a prompt \cite{tessari2025surpassingcosinesimilaritymultidimensional}. This suggests that vector magnitude, which cosine similarity discards, can indeed hold important semantic information.This body of work confirms that the choice of similarity function interacts strongly with the structure of the embedding space.

\subsection{Our Contribution}
The existing literature clearly indicates a move away from the uncritical use of cosine similarity. While significant progress has been made in developing new training objectives and learned metrics, there remains a need for simple, robust, and parameter-free similarity functions that can serve as direct replacements for cosine similarity. Our work contributes to this specific area of research. We introduce and conduct a large-scale empirical evaluation of two such metrics, Overlap Similarity (OS) and Hyperbolic Tangent Similarity (HTS). By testing across a wide array of models and datasets, we aim to determine if these functions, through their unique normalization schemes, offer a more reliable measure of semantic similarity in the anisotropic spaces typical of modern PLMs.

\section{Methodology: From Standard Metrics to Robust Alternatives}

Our study evaluates the efficacy of four distinct, parameter-free similarity functions for comparing a pair of sentence embeddings, denoted as vectors x and y in $R^d$. This section first details the limitations of the two standard baseline metrics—Dot Product and Cosine Similarity—and then introduces Overlap Similarity (OS) and Hyperbolic Tangent Similarity (HTS) as proposed solutions to these identified issues.

\subsection{The Limitations of Standard Baselines}

\paragraph{Dot Product} The dot product is the most fundamental vector comparison, but its utility is hampered by its direct dependence on vector magnitude. A vector's L2-norm can be influenced by factors like sentence length or word frequency, which do not reliably correlate with semantic importance. This creates a scenario where a long, generic sentence could have a high similarity score with another sentence merely due to its large norm, not its semantic content. Its unbounded nature also makes it difficult to interpret and normalize across different models and datasets. The formula is:

$$
\text{sim}_{\text{dot}}(x, y) = x \cdot y = \sum_{i=1}^{d} x_i y_i
$$
 
\paragraph{Cosine Similarity (CS)} To solve the magnitude problem, Cosine Similarity became the standard. By normalizing the dot product, it discards vector magnitudes entirely to focus solely on the angle between vectors. It is defined as:

$$
\text{sim}_{\text{cos}}(x, y) = \frac{x \cdot y}{\|x\|_2 \|y\|_2}
$$
 
However, this approach introduces its own problems. As demonstrated by Gao et al. (2021), embeddings from modern language models are often \textbf{anisotropic}, meaning they occupy a narrow cone in the vector space. When all vectors are clustered together, the angle between any two vectors is small, resulting in high cosine similarity scores for nearly all pairs. This severely reduces the metric's discriminative power. Furthermore, by completely discarding vector magnitude, Cosine Similarity may be throwing away useful information, as the norm could encode a notion of semantic specificity or importance.

\subsection{Proposed Metrics for Robust Similarity}

Our proposed metrics, OS and HTS, are designed to overcome these limitations by incorporating vector magnitude in a more principled and robust manner.

\paragraph{Overlap Similarity (OS)} We propose Overlap Similarity as a metric that intelligently balances vector direction and magnitude. We hypothesize that its unique normalization scheme is more resilient to the effects of anisotropy. It is formally defined as:

$$
\text{sim}_{\text{os}}(x, y) = \frac{x \cdot y}{\|x\|_2^2 + \|y\|_2^2 - |x \cdot y| + \epsilon}
$$
 
\textbf{Intuition:} Unlike Cosine Similarity, which normalizes by the product of individual vector norms, OS normalizes by a term that represents the "total energy" of the two vectors, corrected for their alignment. The denominator, 

$$
{\|x\|_2^2 + \|y\|_2^2 - |x \cdot y| + \epsilon}
$$

is analogous to the principle of inclusion-exclusion. It sums the squared magnitudes and then subtracts their shared component (the absolute dot product), effectively measuring the "union" of the vectors' energies. This makes the normalization factor dependent on the relationship \textit{between} the vectors, not just their independent properties. This relational normalization may provide a more stable and meaningful similarity score when all vectors are already pointing in a similar direction.

\paragraph{Hyperbolic Tangent Similarity (HTS)} This metric introduces a non-linear transformation that we hypothesize is better suited for capturing complex semantic relationships. HTS is defined as:

$$
\text{sim}_{\text{hts}}(x, y) = \tanh\left(2 \cdot \frac{x \cdot y}{\|x\|_2^2 + \|y\|_2^2 + \epsilon}\right)
$$

\textbf{Intuition:} HTS tackles the problem in two ways. First, it uses a simple and stable normalization term: the sum of the squared L2-norms. Second, and more importantly, it applies the tanh function, which "squashes" the resulting value into a fixed [-1,1] range. This nonlinear transformation has two key benefits:
\begin{enumerate}
\item \textbf{Robustness to Outliers:} It makes the metric less sensitive to extreme values. A few unusually large dimensions in the vectors won't disproportionately affect the final similarity score.
\item \textbf{Modeling Non-Linearity:} The relationship between vector similarity and true semantic similarity may not be linear. The S-shaped curve of the tanh function amplifies differences for mid-range similarity scores while compressing differences at the extremes (very similar or very dissimilar pairs). This may better reflect how humans perceive similarity.
\end{enumerate}

By retaining information about vector magnitude and applying more robust normalization and non-linear scaling, we posit that OS and HTS can provide a more accurate and discriminative measure of semantic similarity than their standard counterparts.

\section{Experimental Setup}

To provide a rigorous and reproducible comparison of the similarity metrics, we designed a comprehensive experimental protocol. This section details the motivation for our experimental design, our core evaluation methodology, and the specific models, datasets, and statistical methods used in our study.

\subsection{Motivation for Experimental Design}
The primary goal of our experimental design is to test the hypothesis that OS and HTS are more robust and generalizable similarity metrics than standard baselines. To achieve this, our design is guided by the principle of comprehensive evaluation across multiple axes of variation.
\begin{itemize}
\item \textbf{Model Diversity:} By selecting four sentence-transformer models with varying sizes, architectures, we aim to demonstrate that our findings are not an artifact of a single model's embedding geometry but hold true across a representative sample of modern encoders.
\item \textbf{Task Diversity:} The inclusion of eight datasets spanning Semantic Textual Similarity (STS), paraphrase identification, and Natural Language Inference (NLI) is crucial. This variety allows us to assess whether the proposed metrics excel at just one type of similarity or provide a consistent advantage across different semantic tasks.
\item \textbf{Metric Completeness:} Relying on a single evaluation score can be misleading. We use Mean Squared Error (MSE) to capture the absolute accuracy of the similarity scores and Spearman's Rank Correlation ($\rho$) to evaluate their relative ordering. This dual-metric approach provides a more complete picture of performance.
\item \textbf{Statistical Rigor:} To move beyond anecdotal claims of improvement, we employ formal statistical testing. The Wilcoxon signed-rank test confirms if an improvement is statistically significant, while bootstrapped confidence intervals provide a crucial measure of the magnitude and certainty of the effect.
\end{itemize}

\subsection{Core Evaluation Methodology}
It is crucial to note that our experimental procedure does not involve any model training or fine-tuning. The core of our methodology is a post-hoc analysis of existing, pre-trained sentence embeddings. The process for each experiment is as follows:
\begin{enumerate}
\item \textbf{Embedding Generation:} We take an off-the-shelf, pre-trained sentence-transformer model and use it to generate static, dense vector embeddings for each sentence in a given dataset's evaluation split. This step is performed only once per model-dataset pair.
\item \textbf{Similarity Function Application:} With the static embeddings generated, we then apply each of the four mathematical similarity functions (Dot Product, Cosine, OS, and HTS) to the pairs of vectors. This is a direct, zero-shot application of a mathematical formula.
\end{enumerate}
This parameter-free approach is fundamental to our contribution. We are not proposing a new way to train embeddings but are instead investigating the most effective way to compare the embeddings that these models already produce. This ensures that any observed performance gains are attributable solely to the choice of similarity metric itself, not to any additional training or learned parameters.

\subsection{Models}
We selected four widely-used, publicly available sentence-transformer models to serve as our base encoders. These models were chosen to represent a variety of sizes, architectures, and training paradigms, ensuring that our findings are not specific to a single model class. The models are:
\begin{itemize}
\item \texttt{all-MiniLM-L6-v2} from hugging face \cite{sbert_minilm_2022}
\item \texttt{paraphrase-mpnet-base-v2} from hugging face \cite{sbert_paraphrase_mpnet_2021}
\item \texttt{all-mpnet-base-v2} from hugging face \cite{sbert_all_mpnet_2021}
\item \texttt{BAAI/bge-large-en-v1.5} from hugging face \cite{bge_embedding}
\end{itemize}

\subsection{Datasets}
Our evaluation is conducted on a carefully curated suite of seven benchmark datasets. This collection was intentionally selected to provide a comprehensive and challenging assessment of semantic understanding. These datasets are standard in the NLP community, ensuring our results are directly comparable with prior and future work. Their diversity across three distinct tasks—Semantic Textual Similarity (STS), paraphrase identification, and Natural Language Inference (NLI)—allows us to rigorously test the generalizability and robustness of each similarity metric. For all datasets, gold-standard scores were normalized to a [0,1] range. The datasets are:

\begin{itemize}
\item \textbf{Semantic Textual Similarity:} GLUE-STSB \cite{wang2019gluemultitaskbenchmarkanalysis}, SICK \cite{marelli-etal-2014-sick}, and STS16 \cite{agirre-etal-2016-semeval}.
\item \textbf{Paraphrase Identification:} PAWS \cite{zhang-etal-2019-paws} and Quora Question-Pairs \cite{quora-question-pairs}.
\item \textbf{Natural Language Inference:} SNLI \cite{bowman2015largeannotatedcorpuslearning} and MultiNLI \cite{williams-etal-2018-broad}. For NLI tasks, we frame the problem as a similarity task by mapping the entailment label to a score of 1.0 and all other labels to 0.0.
\end{itemize}
All datasets were accessed via the Hugging Face datasets library, and we used the standard labeled evaluation splits (e.g., validation or test) as recommended by community best practices.

\subsection{Evaluation Protocol and Metrics}
For each model and dataset pair, we first generate embeddings for all sentence pairs. We then compute similarity scores using each of the four metrics. To evaluate the quality of these scores, we employ two distinct metrics:

\paragraph{Mean Squared Error (MSE)} measures the average squared difference between the predicted similarity scores and the ground-truth labels. It provides a measure of absolute error, penalizing large deviations more heavily. A lower MSE indicates better performance.

\paragraph{Spearman's Rank Correlation ($\rho$)} assesses the monotonic relationship between the ranks of the predicted scores and the ranks of the gold labels. This metric is insensitive to the absolute scale of the scores and instead evaluates whether a metric correctly orders the sentence pairs from least to most similar. A higher $\rho$ value (closer to 1) indicates better performance.

By using both MSE and Spearman's $\rho$, we gain a comprehensive view of performance, capturing both the accuracy of the absolute scores and the correctness of the relative rankings.

\subsection{Statistical Analysis}
To ensure the reliability of our conclusions, we perform rigorous statistical significance testing.
\begin{itemize}
\item \textbf{Wilcoxon Signed-Rank Test:} We use a one-sided Wilcoxon signed-rank test to compare the distributions of squared errors produced by our proposed metrics against the baselines. This non-parametric test allows us to determine if the observed improvements are statistically significant (p $<$ 0.05).
\item \textbf{Bootstrapped Confidence Intervals:} To quantify the magnitude of the performance difference, we employ bootstrapping to calculate the 95\% confidence interval for the mean difference in MSE between metrics. A confidence interval that is strictly positive indicates with high confidence that the proposed metric is superior.
\end{itemize}
This two-pronged approach allows us not only to state that one metric is better, but also by how much.

\subsection{Implementation and Reproducibility Details}
To ensure full reproducibility in accordance with AAAI guidelines, we detail our implementation and experimental parameters below.

\paragraph{Computing Infrastructure} All experiments were conducted on a workstation equipped with an NVIDIA RTX 4070 GPU with 6GB of VRAM. The operating system was Windows. Key software libraries include Python 3.8.19, PyTorch 2.4.1 (+cu121), Sentence-Transformers 3.2.1, Datasets 3.1.0, Pandas 2.0.3, Scikit-learn 1.3.2, and Scipy 1.10.0.

\paragraph{Hyperparameters and Pre-processing} A key advantage of our methodology is that the core similarity metrics (OS and HTS) are parameter-free, requiring no hyperparameter tuning. The only pre-processing step involves normalizing the gold-standard scores of some datasets to a [0, 1] range, which is handled by a simple, self-contained lambda function within our evaluation script.

\paragraph{Code Availability} The complete source code used to conduct all experiments and generate all results reported in this paper will be made publicly available upon publication under an MIT license. The code is extensively commented to detail the implementation of each step and is provided in the appendix.

\paragraph{Randomness and Number of Runs} The core evaluation process—generating embeddings and applying similarity functions—is deterministic. Therefore, each reported MSE and Spearman correlation score is the result of a single, reproducible run. The only stochastic component of our analysis is the bootstrapping procedure for calculating confidence intervals, for which we used 1000 resamples. This procedure relies on the standard pseudo-random number generator in the NumPy library, which was seeded for reproducibility of the statistical analysis itself.

\section{Results}

Our experiments consistently demonstrate the superior performance of Overlap Similarity (OS) and Hyperbolic Tangent Similarity (HTS) over standard baselines in classification based tasks. To illustrate the robustness of our findings, this section presents the detailed results for all the four embedding models.

For each model, we present the Mean Squared Error (MSE) and Spearman's Rank Correlation ($\rho$). The full statistical analysis, including p-values and 95\% confidence intervals for all model-dataset pairs, is provided in the Technical Appendix.

\subsection{Model: all-mpnet-base-v2}

The results for the \texttt{all-mpnet-base-v2} model are shown in Table \ref{tab:mse_mpnet} and Table \ref{tab:spearman_mpnet}. On all classification-based tasks (Quora, PAWS, SNLI, etc.), OS and HTS achieve a lower MSE. Notably, on the fine-grained STS tasks (GLUE-STSB, SICK), the standard baselines remain competitive or superior, highlighting a key interaction between the metric, model, and task type.

\begin{table}[h!]
\centering
\begin{tabular}{lcccc}
\toprule
\textbf{Dataset} & Dot & Cosine & OS & HTS \\
\midrule

GLUE-STSB & 0.1916 & 0.1916 & \textbf{0.1732} & 0.1875 \\
SICK & \textbf{0.0316} & \textbf{0.0316} & 0.0773 & 0.0490 \\
\midrule
Quora & 0.2487 & 0.2487 & \textbf{0.1773*} & 0.2067* \\
PAWS & 0.5109 & 0.5109 & 0.4765* & \textbf{0.3347*} \\
SNLI & 0.1872 & 0.1872 & \textbf{0.1627*} & 0.1791* \\
MultiNLI & 0.2295 & 0.2295 & \textbf{0.1804*} & 0.2102* \\
STS16 & 0.0593 & 0.0593 & \textbf{0.0414*} & 0.0533* \\

\bottomrule
\end{tabular}
\caption{ Mean Squared Error (MSE) using all-mpnet-base-v2 as the embedding model. Lower is better. Bold indicates the best metric. * denotes a statistically significant improvement over both Cosine Similarity and Dot Product (p $<$ 0.05).}
\label{tab:mse_mpnet}
\end{table}

\begin{table}[h!]
\centering
\begin{tabular}{lcccc}
\toprule
\textbf{Dataset} & Dot & Cosine & OS & HTS \\
\midrule
GLUE-STSB & 0.4282 & 0.4282 & 0.4282 & 0.4282 \\
SICK & 0.8049 & 0.8049 & 0.8049 & 0.8049 \\
\midrule
Quora & 0.6484 & 0.6484 & 0.6484 & 0.6484 \\
PAWS & 0.2616 & 0.2616 & 0.2616 & 0.2616 \\
SNLI & 0.5468 & 0.5468 & 0.5468 & 0.5468 \\
MultiNLI & 0.4994 & 0.4994 & 0.4994 & 0.4994 \\
STS16 & 0.8003 & 0.8003 & 0.8003 & 0.8003 \\
\bottomrule
\end{tabular}
\caption{Spearman's $\rho$ using all-mpnet-base-v2 as the embedding model. Higher is better. Notice that the values are equal for all metrics}
\label{tab:spearman_mpnet}
\end{table}

\subsection{Model: all-MiniLM-L6-v2}
For the smaller \texttt{all-MiniLM-L6-v2} model, we observe a similar pattern of results, as shown in Table \ref{tab:mse_minilm} and \ref{tab:spearman_minilm}. The proposed metrics again demonstrate a consistent advantage on classification-style tasks.

\begin{table}[h!]
\centering
\begin{tabular}{lcccc}
\toprule
\textbf{Dataset} & Dot & Cosine & OS & HTS \\
\midrule
GLUE-STSB & 0.1961 & 0.1961 & \textbf{0.1822} & 0.1919 \\
SICK & \textbf{0.0220} & \textbf{0.0220} & 0.0532 & 0.0358 \\
\midrule
Quora & 0.2639 & 0.2639 & \textbf{0.1910*} & 0.2169* \\
PAWS & 0.5179 & 0.5179 & 0.4880* & \textbf{0.3372*} \\
SNLI & 0.1989 & 0.1989 & \textbf{0.1716*} & 0.1882* \\
MultiNLI & 0.2646 & 0.2646 & \textbf{0.2071*} & 0.2331* \\
STS16 & 0.0656 & 0.0656 & \textbf{0.0441*} & 0.0567* \\
\bottomrule
\end{tabular}
\caption{Mean Squared Error (MSE) using all-MiniLM-L6-v2 as the embedding model. Lower is better. * denotes a statistically significant improvement over both Cosine Similarity and Dot Product (p $<$ 0.05).}
\label{tab:mse_minilm}
\end{table}

\begin{table}[h!]
\centering
\begin{tabular}{lcccc}
\toprule
\textbf{Dataset} & Dot & Cosine & OS & HTS \\
\midrule
GLUE-STSB & 0.3739 & 0.3739 & 0.3739 & 0.3739 \\
SICK & 0.7715 & 0.7715 & 0.7715 & 0.7715 \\
\midrule
Quora & 0.6094 & 0.6094 & 0.6094 & 0.6094 \\
PAWS & 0.2250 & 0.2250 & 0.2250 & 0.2250 \\
SNLI & 0.5038 & 0.5038 & 0.5038 & 0.5038 \\
MultiNLI & 0.3901 & 0.3901 & 0.3901 & 0.3901 \\
STS16 & 0.7899 & 0.7899 & 0.7899 & 0.7899 \\
\bottomrule
\end{tabular}
\caption{Spearman's $\rho$ using all-MiniLM-L6-v2 as the embedding model. Higher is better. Notice that the values are equal for all metrics}
\label{tab:spearman_minilm}
\end{table}

\subsection{Model: BAAI/bge-large-en-v1.5}
With the larger \texttt{bge-large-en-v1.5} model too, the trend continues (Tables \ref{tab:mse_bge} and \ref{tab:spearman_bge}), reinforcing the generalizability of our findings across models of different scales. The performance gains for OS and HTS on classification tasks are particularly pronounced with this powerful model.

\begin{table}[h!]
\centering
\begin{tabular}{lcccc}
\toprule
\textbf{Dataset} & Dot & Cosine & OS & HTS \\
\midrule
GLUE-STSB & 0.2044 & 0.2044 & \textbf{0.1698} & 0.1916 \\
SICK & \textbf{0.0139} & \textbf{0.0139} & 0.0199 & 0.0219 \\
\midrule
Quora & 0.3321 & 0.3321 & \textbf{0.2350*} & 0.2614* \\
PAWS & 0.5243 & 0.5243 & 0.4968* & \textbf{0.3393*} \\
SNLI & 0.2820 & 0.2820 & \textbf{0.2030*} & 0.2422* \\
MultiNLI & 0.3062 & 0.3062 & \textbf{0.2141*} & 0.2571* \\
STS16 & 0.1271 & 0.1271 & \textbf{0.0625*} & 0.0931* \\
\bottomrule
\end{tabular}
\caption{Mean Squared Error (MSE) using BAAI/bge-large-en-v1.5 as the embedding model. Lower is better. * denotes a statistically significant improvement over both Cosine Similarity and Dot Product (p $<$ 0.05).}
\label{tab:mse_bge}
\end{table}

\begin{table}[h!]
\centering
\begin{tabular}{lcccc}
\toprule
\textbf{Dataset} & Dot & Cosine & OS & HTS \\
\midrule
GLUE-STSB & 0.5191 & 0.5191 & 0.5191 & 0.5191 \\
SICK & 0.8123 & 0.8123 & 0.8123 & 0.8123 \\
\midrule
Quora & 0.6309 & 0.6309 & 0.6309 & 0.6309 \\
PAWS & 0.2443 & 0.2443 & 0.2443 & 0.2443 \\
SNLI & 0.5259 & 0.5259 & 0.5259 & 0.5259 \\
MultiNLI & 0.5742 & 0.5742 & 0.5742 & 0.5742 \\
STS16 & 0.8649 & 0.8649 & 0.8649 & 0.8649 \\
\bottomrule
\end{tabular}
\caption{Spearman's $\rho$ using BAAI/bge-large-en-v1.5 as the embedding model. Higher is better. Notice that the values are equal for all metrics.}
\label{tab:spearman_bge}
\end{table}

\subsection{Model: paraphrase-mpnet-base-v2}
With \texttt{paraphrase-mpnet-base-v2} model too, we see the same trend (Tables \ref{tab:mse_para} and \ref{tab:spearman_para}), reinforcing the generalizability of our findings across models.

\begin{table}[h!]
\centering
\begin{tabular}{lcccc}
\toprule
\textbf{Dataset} & Dot & Cosine & OS & HTS \\
\midrule
GLUE-STSB & 37.5341 & 0.0329 & \textbf{0.0180} & 0.0254 \\
SICK & 48.9023 & \textbf{0.0232} & 0.0500 & 0.0353 \\
\midrule
Quora & 56.8058 & 0.2745 & {0.1962*} & 0.2222* \\
PAWS & 79.4205 & 0.5232 & 0.4955* & {0.3387*} \\
SNLI & 38.4963 & 0.2290 & {0.1821*} & 0.2043* \\
MultiNLI & 36.5549 & 0.2174 & {0.1651*} & 0.1993* \\
STS16 & 44.4631 & 0.0684 & {0.0372*} & 0.0573* \\
\bottomrule
\end{tabular}
\caption{Mean Squared Error (MSE) using paraphrase-mpnet-base-v2 as the embedding model. Lower is better. * denotes a statistically significant improvement over both Cosine Similarity and Dot Product (p $<$ 0.05). Note that MSE for dot product is very high, because paraphrase-mpnet-base-v2 produces embeddings with very large magnitudes.}
\label{tab:mse_para}
\end{table}

\begin{table}[h!]
\centering
\begin{tabular}{lcccc}
\toprule
\textbf{Dataset} & Dot & Cosine & OS & HTS \\
\midrule
GLUE-STSB & 0.8737 & 0.8942 & 0.8942 & 0.8942 \\
SICK & 0.7653 & 0.7791 & 0.7790 & 0.7790 \\
\midrule
Quora & 0.6078 & 0.6373 & 0.6372 & 0.6372 \\
PAWS & 0.0931 & 0.2549 & 0.2548 & 0.2548 \\
SNLI & 0.4796 & 0.4842 & 0.4840 & 0.4840 \\
MultiNLI & 0.5620 & 0.5814 & 0.5805 & 0.5805 \\
STS16 & 0.8102 & 0.8402 & 0.8403 & 0.8403 \\
\bottomrule
\end{tabular}
\caption{Spearman's $\rho$ using paraphrase-mpnet-base-v2 as the embedding model. Higher is better. Notice that the values are equal for all metrics.}
\label{tab:spearman_para}
\end{table}

\section{Discussion}
First, we see that for all the embedding models, we see the exact same pattern,  Overlap Similarity (OS) and Hyperbolic Tangent Similarity (HTS) outperm the baseline metrics on all datasets, except GLUE and SICK. The possible reasons are discussed below. Second, the new metrics i.e. OS and HTS have the exact same spearman correlation as the cosine. This means, the sentence pair ranks are the same, but there are improvements in the MSE. The findings indicate that for a majority of common NLP tasks, Overlap Similarity (OS) and Hyperbolic Tangent Similarity (HTS) offer a significant advantage over standard baselines. This section provides an interpretation of these results, discusses their practical implications, and outlines the limitations of the study.

\subsection{Interpreting the Performance Dichotomy}

The most salient finding is the performance difference between classification-based tasks, which typically use more general language, and the fine-grained regression tasks, which are designed to be linguistically nuanced.

\paragraph{Why OS and HTS Excel on General Language Tasks}
On datasets like Quora, SNLI, and PAWS, the sentence pairs are drawn from natural, "in-the-wild" sources. The task is fundamentally to create a decision boundary to separate discrete classes (e.g., 'is duplicate' vs. 'not duplicate', entailment vs. contradiction). We hypothesize that OS and HTS excel here precisely because they incorporate vector magnitude in a controlled way, which helps to better separate these broader categories.
\begin{itemize}
\item \textbf{Overlap Similarity's} relational normalization, which accounts for the "total energy" of the vector pair, may be more effective at separating distinct clusters of points in the embedding space.
\item \textbf{HTS's} non-linear tanh function squashes scores into a fixed range. This S-shaped curve can amplify small differences around the decision boundary, effectively making the separation between classes more distinct, which is ideal for classification.
\end{itemize}

\paragraph{Explaining the Anomaly on Nuanced STS Tasks}
Conversely, on the SICK and GLUE-STSB datasets, the goal is not to classify but to predict a continuous similarity score. These datasets were specifically constructed by linguists to contain very fine-grained and nuanced semantic differences (e.g., "A man is playing a guitar" vs. "A man is playing a musical instrument"). In this context, the standard baselines, particularly Cosine Similarity, were often superior. We posit two primary reasons for this:
\begin{enumerate}
\item \textbf{Noisy Magnitude Information:} It is possible that for these subtle linguistic variations, the semantic signal is almost entirely encoded in the direction of the embedding vector, while the vector's magnitude does not carry useful information and may even represent noise. By completely discarding magnitude, Cosine Similarity inadvertently isolates the cleaner, angular signal. OS and HTS, by attempting to incorporate this potentially noisy magnitude information, are penalized.
\item \textbf{Uniform Vector Norms:} The identical Spearman scores observed across all metrics for several model-dataset pairs suggest that the embeddings for these datasets have highly uniform L2 norms. When all vector magnitudes are nearly equal, the normalization factors of all four metrics become monotonic transformations of each other. This means they will all produce the same ranking of pairs (hence the identical Spearman scores), but their different scaling properties will lead to different absolute errors (MSE).
\end{enumerate}

\subsection{Practical Implications}
The primary implication of our work is practical and immediate. For a wide range of common NLP applications centered on paraphrase detection, semantic search, and inference, practitioners can achieve a significant, zero-cost performance improvement by replacing Cosine Similarity with Overlap Similarity or Hyperbolic Tangent Similarity. As these metrics are parameter-free and computationally inexpensive, they represent a "drop-in" enhancement that requires no additional model training or complex implementation.

\subsection{Limitations and Future Work}
This study, while comprehensive, has several limitations that open avenues for future research. Our analysis was confined to English-language sentence embeddings from BERT-style transformer architectures. Future work should explore whether these findings generalize to other languages and different model architectures (e.g., RNNs, large language models).

Furthermore, the interaction between embedding space geometry and metric performance warrants deeper investigation. A promising direction would be to formally measure the anisotropy of a given model's embeddings and use that to predict which similarity metric will perform best, potentially leading to an adaptive, model-aware similarity function.

\section{Conclusion}

In this paper, we challenged the default use of Cosine Similarity for comparing sentence embeddings, a practice that can be suboptimal in the anisotropic spaces produced by modern language models. Through a large-scale empirical study across multiple models and a diverse set of tasks, we demonstrated that alternative, parameter-free metrics—namely Overlap Similarity (OS) and Hyperbolic Tangent Similarity (HTS)—offer significant performance gains on classification-based tasks such as paraphrase identification and natural language inference.

Our findings also revealed a critical nuance: on fine-grained semantic textual similarity tasks, the standard Cosine Similarity remains a strong baseline, suggesting that the optimal choice of metric is task-dependent. The primary contribution of this work is the identification and validation of OS and HTS as superior, zero-cost, drop-in replacements for Cosine Similarity for a majority of common NLP applications. This provides practitioners with a simple and effective method to improve the performance of their embedding-based systems without the need for additional training.

\bibliography{aaai2026}

\end{document}